# Mining of health and disease events on Twitter: validating search protocols within the setting of Indonesia

Aditya L. Ramadona[a,b*], Rendra Agusta[c], Sulistyawati[d], Lutfan Lazuardi[e], Anwar D. Cahyono[f], Åsa Holmner[g], Fatwa S.T. Dewi[h], Hari Kusnanto[i], Joacim Rocklöv[a,j]

[a]*Department of Public Health and Clinical Medicine, Epidemiology and Global Health, Umeå University, Umeå 90187, Sweden*
[b]*Center for Environmental Studies, Universitas Gadjah Mada, Yogyakarta 55281, Indonesia*
[c]*Department of Psychology, Universitas Mercu Buana, Yogyakarta 55753, Indonesia*
[d]*Department of Public Health, Universitas Ahmad Dahlan, Yogyakarta 55164, Indonesia*
[e]*Department of Health Policy and Management, Faculty of Medicine, Universitas Gadjah Mada, Yogyakarta 55281, Indonesia*
[f]*District Health Office, Yogyakarta 55165, Indonesia*
[g]*Department of Radiation Sciences, Umeå University, Umeå 90187, Sweden*
[h]*Department of Health Behavior, Environment and Social Medicine, Faculty of Medicine, Universitas Gadjah Mada, Yogyakarta 55281, Indonesia*
[i]*Department of Community and Family Medicine, Faculty of Medicine, Universitas Gadjah Mada, Yogyakarta 55281, Indonesia*
[j]*Heidelberg University Medical School, Institute of Public Health, Heidelberg 69120, Germany*

**Abstract**

This study seeks to validate a search protocol of ill health-related terms using Twitter data which can later be used to understand if, and how, Twitter can reveal information on the current health situation. We extracted conversations related to health and disease postings on Twitter using a set of pre-defined keywords, assessed the prevalence, frequency, and timing of such content in these conversations, and validated how this search protocol was able to detect relevant disease tweets. Classification and Regression Trees (CART) algorithm was used to train and test search protocols of disease and health hits comparing to those identified by our team. The accuracy of predictions showed a good validity with AUC beyond 0.8. Our study shows that monitoring of public sentiment on Twitter can be used as a real-time proxy for health events.

## 1. Introduction

Twitter is a free social networking and micro-blogging service that enables its users to read and share messages no longer than 140-characters. As of May 2016, there are 24.34 million Indonesian, or around 10% of the population being active monthly on Twitter [1], sharing news, events, as well as their personal feelings and experiences including health-related information. Twitter offers a potential for data mining of public information flows [2] and these massive data sources may be exploited for public health monitoring and surveillance purposes [3].

Previous studies have explored the use of Twitter, for example, to track levels of disease activity [4], to predicts heart disease mortality [5], and for measuring health-related quality of life [6]. However, the validity of twitter mining protocols to correctly detect health and disease events is one methodological challenge of this media. This study seeks to validate a search protocol of ill health-related terms using real-time Twitter data which can later be used to understand if, and how, twitter can reveal information on the current health situation in Indonesia.

In this validation study of mining protocols, we: 1) extracted geo-located conversations related to health and disease postings on Twitter using a set of pre-defined keywords, 2) assessed the prevalence, frequency and timing of such content in these conversations, and 3) validated how this search protocol was able to detect relevant disease tweets.

## 2. Materials and Methods

We started by developing groups of words and phrases relevant to disease symptoms and health outcomes to extract relevant content from Twitter Search API [7]. Since this study is intended to reveal the relationships in the setting of Indonesia, we only considered and filtered tweets in Bahasa by using the following keywords.

"rumah OR sakit OR rawat OR inap OR demam OR panas -cuaca OR berdarah OR pendarahan OR tombosit OR badan OR muntah OR badan OR tua OR ':("'.

We removed retweets and, therefore, only collected 390 tweets in Bahasa. We obtained 1,632 unique words from such tweets after removing punctuation, numbers, capitalization, and the Bahasa stop-words (e.g., kamu and aja). Many words appeared infrequently and we considered only the 22 highest words frequencies (i.e., words that appear at least 10 times, Fig. 1).

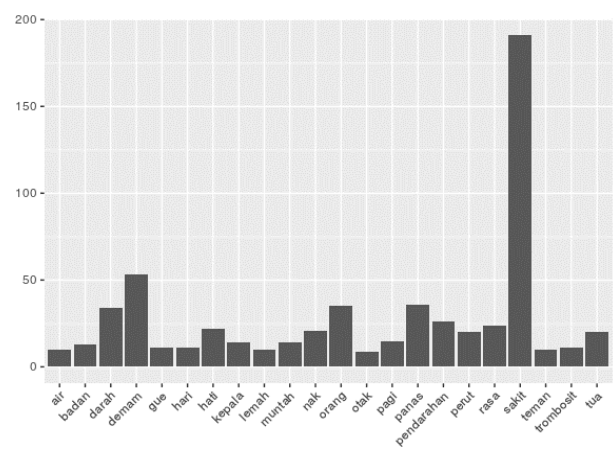

Fig. 1. Words that appear more than 10 times

We developed a model of twitter search term protocol hits to our team assigned positive or neutral tweets content in respect to health events. We used the rpart package [8], in R software environment [9], to implement the CART algorithm [10]. Predictors were generated from the 22 words with highest frequency, i.e. by counting the number of such words in a tweet. The model was trained and validated by dividing the 390 tweets from the historical feeds from Twitter search API randomly, into two datasets were 273 tweets (70%) was used for training, and 117 tweets (30%) was used for validating the model.

For further investigation, we tested the model to the Twitter real-time feeds collecting a total of 1,145,649 tweets using Twitter stream API [2] by recording conversations within Indonesia region (between latitudes 11°S and 6°N, and longitudes 95°E and 141°E), for one week (26th July – 1st August 2016). For this, we selected randomly 100 True and 100 False predictions from all hits to be further checked for accuracy.

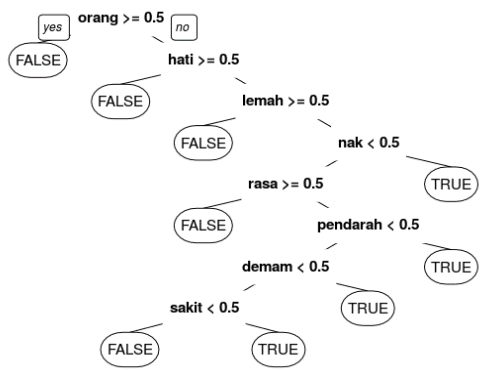

Fig. 2. Classification and regression trees

To evaluate the feasibility of quantifying and using this protocol to detect disease trends, we calculate its performance in term of sensitivity, specifity, positive predictive value (PPV), and negative predictive value (NPV).

## 3. Results and Discussion

We found that hati (heart), rasa (feel) and perut (stomach) are the most highly correlated words with sakit (sick, ill, pain). Other words showed a lower correlation to sakit (correlation<0.1). We further explored the classification and regression trees model (Fig. 2) and found that if the word orang (person), hati, lemah (weak), or rasa are in a tweet, then the model predicted FALSE more often. On the contrary, if the word nak (son, daughter), pendarahan (hemorrhage, bleeding), demam (fever), or sakit are in a tweet, then it predicts TRUE more often. This tree makes sense intuitively since orang, hati, lemah, and rasa are generally seen as words represent the state of broken heart (sakit hati), while nak, pendarahan, demam, or sakit are words represents a state of pain and illness.

Furthermore, we were also able to extract geo-located conversations related to disease symptoms and health outcomes, as well as its prevalence, frequency, and timing, by aggregating daily geo-located tweets. Figure 3 shows 6,109 tweets of predicted hits from the model, within Indonesia region during July 26th - August 1st, 2016. We found that most of health or disease risk signal appeared in the western part of Indonesia, e.g. Sumatera and Java Island.

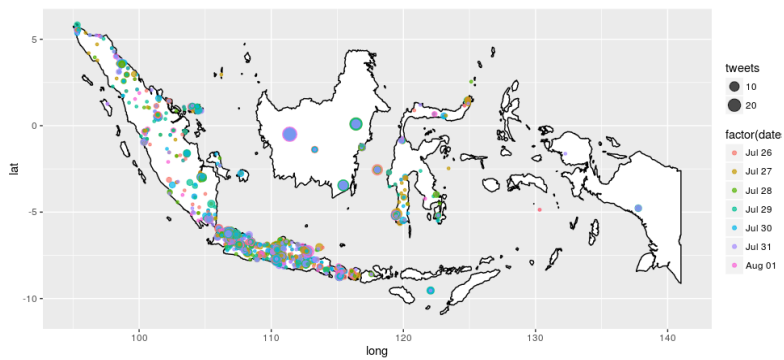

Fig. 3. Geo-located conversations related to health (or disease)

As was to be expected, the majority of tweets came from Java island, the most densely populated island in Indonesia. Other areas included some of the capital of districts and provinces in Sumatra, Kalimantan and Sulawesi (Fig. 3). The availability of telecommunications infrastructure and the characteristics of the people who live in a region is likely to influence the number of Twitter users in such region. Although this study did not examine and compare the distribution of the number of Twitter users in each region of Indonesia, the methods developed in this study has been able to detect the signal, i.e. the tweet containing a health risk information, in nearly real-time.

The accuracy of the prediction (Table 1) to categorize the extracted tweets appear good, both in validation and testing dataset (AUC, 0.82, 0.70). However, we found that the sensitivity and NPV were decreased, meanwhile, specifity and PPV were increased (Table 1).

The predictions based on the tweets contain information that refers to someone whose being unhealthy, and therefore, we only detected 6,109 TRUE results of 1,145,649 tweets from Twitter real-time feeds. This means the number of tweets that may contain the conversations related to disease symptoms and health outcomes was very small (0.5%) compared to the entire tweets stream. We believe its unbalanced dataset may contribute in decreasing the ability of the model to correctly classify a tweet. Therefore, in subsequent research, it is necessary to investigate and compare the current results with a more robust algorithm in term of dealing with an unbalanced dataset (e.g., anomaly detection algorithm).

Table 1. Model Accuracy

|  | Validation | Testing |
| --- | --- | --- |
| Sensitivity | 80.0 | 42.0 |
| Specifity | 84.6 | 98.0 |
| Positive Predictive Value | 86.7 | 95.5 |
| Negative Predictive Value | 77.2 | 62.8 |
| AUC | 0.82 | 0.70 |

* Corresponding author. Tel.:+62 8522 9394 370.
Email: alramadona@ugm.ac.id, alramadona@outlook.com

A number of limitations are apparent for this study, especially, for example, the team member involved in building the dataset to train the model were mainly from academics and speakers of local languages used in the western region of Indonesia. Considering the complexity of how people express ideas on Twitter, especially the nature of Indonesians who has a broad variety of local languages [11], it is necessary to involve additional team member from another discipline, such as Linguistic, as well as health workers from the eastern region of Indonesia in order to make a more robust validation of the mining protocol.

Previous studies have shown that the implementation of disease control programs has not given satisfactory result due to, among others, surveillance program implementation in Indonesia which are mostly passive [12], and the different levels of public knowledge about a particular disease in the community [13]. Therefore, despite some of the limitations mentioned above, this case study highlight the wealth of information in Twitter conversation that might be utilized to develop an early warning to identify disease trend, as a part to enhance the surveillance system has been running. The models allowed some recognition, although not perfect, of health risk signal at nearly real-time. Thus, development of early warning system could benefit from these predictions since one of the main reason for disease control being less successful is related to the poor and reactive management targeting interventions too late in the epidemics [14]. It may target high-risk people and periods, for when health education and public health interventions can effectively prepare communities and potentially curb the epidemic.

## 4. Conclusion

In this study, we validated a search protocol of health-related terms using real-time Twitter data within the setting of Indonesia. Our study shows that monitoring of public sentiment on Twitter, combined with contextual knowledge about the disease, can detect health and disease tweets and potentially be used as a valuable nearly real-time proxy for health-related indicators. Furthermore, social media analysis might be used to identify a user or community with increased health risks over a specific spatial and temporal range.

**Acknowledgements**

The authors gratefully acknowledge use of Azure cloud computing platform, sponsored by Microsoft.